\pdfoutput=1

\documentclass[11pt]{article}

\usepackage[table]{xcolor} 

\usepackage[preprint]{acl}

\usepackage{times}
\usepackage{latexsym}

\usepackage[T1]{fontenc}

\usepackage[utf8]{inputenc}

\usepackage{microtype}

\usepackage{inconsolata}

\usepackage{graphicx}

\usepackage[utf8]{inputenc} 
\usepackage[T1]{fontenc}    
\usepackage{hyperref}       
\usepackage{url}            
\usepackage{booktabs}       
\usepackage{amsfonts}       
\usepackage{nicefrac}       
\usepackage{microtype}      
\usepackage{xcolor}         
\usepackage{graphicx}
\usepackage{algorithmic}
\usepackage{multirow}
\usepackage[linesnumbered,ruled,vlined]{algorithm2e}
\usepackage{amssymb}
\usepackage{cite}
\usepackage{amsmath}

\usepackage{color}
\usepackage{colortbl}      
\usepackage[most]{tcolorbox}

\usepackage{subcaption}
\usepackage{makecell}
\usepackage[symbol]{footmisc}
\usepackage{fancyhdr}
\usepackage{hyperref}

\definecolor{myred}{HTML}{C00000}
\definecolor{myblue}{HTML}{002060}

\title{Do MLLMs Really See It: Reinforcing Visual Attention in Multimodal LLMs}

\author{Siqu Ou \textsuperscript{\rm 1,2,}\thanks{Work done during an internship at TeleAI} \quad
  Tianrui Wan \textsuperscript{\rm 1,3} \quad
  Zhiyuan Zhao  \textsuperscript{\rm 1,}\thanks{Corresponding author} \quad
  Junyu Gao  \textsuperscript{\rm 1,3}  \\
  {\bf Xuelong Li  \textsuperscript{\rm 1,}}\thanks{Corresponding author} \\
    \textsuperscript{\rm 1} TeleAI \quad
    \textsuperscript{\rm 2} Shanghai Jiao Tong University \quad
    \textsuperscript{\rm 3} Northwestern Polytechnical University
  }

\begin{document}
\maketitle

\begin{abstract}
While chain-of-thought (CoT) reasoning has substantially improved multimodal large language models (MLLMs) on complex reasoning tasks, existing approaches largely rely on long textual reasoning trajectories and provide limited mechanisms for learning stable visual attention policies. Our analysis shows that current MLLMs exhibit weak visual focus: early-stage visual misalignment is rarely corrected during subsequent reasoning, leading to error propagation and failed inferences. We argue that this limitation stems from inadequate credit assignment for visual attention during training. To address this issue, we propose SAYO, a visual reasoning model trained with a reinforcement learning (RL) framework that introduces a region-level visual attention–based reward. This reward explicitly aligns optimization signals with visually grounded reasoning steps, enabling the model to learn more reliable attention behaviors. Extensive experiments across multiple multimodal benchmarks demonstrate that SAYO consistently improves performance on diverse reasoning and perception tasks. Project is open at \url{https://github.com/Cratileo/Sayo}.
\end{abstract}
\section{Introduction}

\begin{figure}[t]
    \centering
    \includegraphics[width=.95\linewidth]{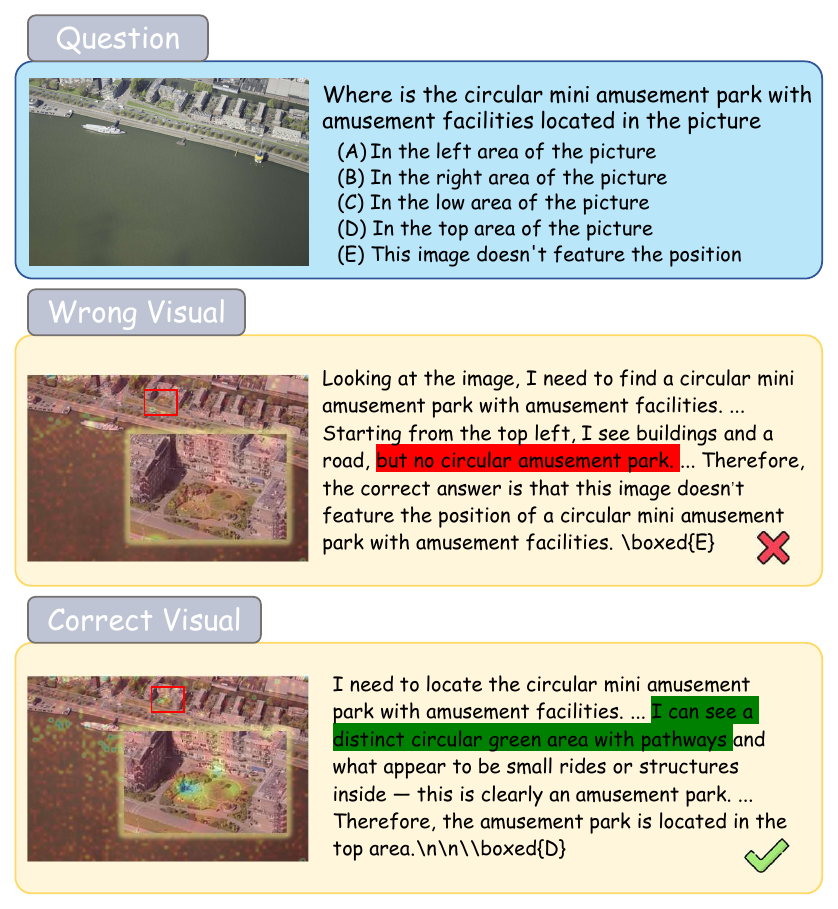}
    \caption{In the reasoning process of CoT, initial visual focusing errors can mislead the inference processing. We emphasize enhancing the model's visual capabilities to ensure MLLMs can proactively focus on text-relevant visual regions. The resolution of image is 2752x1824. The attention map displays the average attention of all generated tokens.}
    \label{fig:first}
\end{figure}

Recent advances in multimodal large language models (MLLMs)~\citep{qwen3vl, internvl3.5, step3vl} have significantly improved performance on complex reasoning tasks that require integrating visual and linguistic information. In particular, chain-of-thought (CoT) reasoning has emerged as an effective mechanism for decomposing complex problems into sequential inference steps~\citep{mmreact,llavacot}. Despite these successes, existing MLLMs continue to exhibit fundamental limitations when reasoning over visually complex inputs, especially in scenarios requiring precise and sustained visual grounding across long reasoning trajectories. Most prior approaches emphasize textual reasoning processes or rely on heuristic visual prompt engineering, such as programmatic region highlighting or prompt reflection mechanisms~\citep{lookback}. While these methods can indirectly influence visual perception, they do not explicitly address how visual attention behaviors are learned during training. Prior works have begun to probe this problem that current MLLMs often develop unstable visual attention policies, particularly when confronted with complex multi-object scenes or information-dense documents~\citep{liu2025seeingbelievingprobingdisconnect,yang2025learninglookdisentangledcurriculum}. These findings suggest that MLLMs may struggle to focus their visual attention and become distracted by an excess of other visual signals. Although techniques such as ViP~\citep{vip} introduce visual cues to mark regions of interest, their effectiveness ultimately depends on the model’s pre-existing attention behavior, which remains insufficiently optimized.

Previous studies~\citep{tong2024eyeswideshutexploring,verma2024crossmodalprojectionmultimodalllms} have revealed a consistent pattern: when an MLLM attends to incorrect visual regions at early inference stages, this misalignment is rarely corrected during subsequent reasoning. Instead, erroneous visual assumptions propagate through the chain of thought, leading to systematic inference failures. This phenomenon is especially pronounced in long reasoning sequences, where early errors exert a disproportionate influence on final predictions. We argue that this limitation reflects a broader optimization deficiency, rather than a lack of representational capacity. Specifically, existing training objectives fail to provide effective credit assignment signals for learning reliable visual attention behaviors during multimodal reasoning.

To investigate this issue, we evaluate representative MLLMs on challenging visual question answering benchmarks involving complex multi-object scenes. As shown in Figure\ref{fig:first}, analysis indicates that incorrect predictions are frequently accompanied by misplaced attention or inaccurate spatial localization. Quantitatively, we observe a strong correlation between visual attention accuracy and overall task performance, supporting the hypothesis that attention misalignment is a primary driver of reasoning errors.

Motivated by these findings, we propose Entropy-Based Target Attention Reward, a novel reinforcement learning–based framework designed to improve visual information learning in MLLMs. Rather than relying on external visual prompts or tools, it introduces an attention-aware reward that directly aligns optimization signals with visually grounded reasoning steps. The reward is selectively applied based on token-level entropy, encouraging the model to prioritize visual information at uncertain decision points. This design provides a principled mechanism for addressing credit assignment in multimodal reasoning. As illustrated in Figure~\ref{fig:first}, models trained with this reward exhibit significantly improved attention alignment during inference, consistently focusing on task-relevant visual regions at critical reasoning steps. Importantly, this improvement emerges without requiring explicit visual prompts, special tokens, or iterative prompt refinement at inference time. The resulting model is able to maintain stable visual grounding throughout long reasoning trajectories.

In summary, this paper makes the following contributions:
\begin{itemize}
    \item \textbf{An optimization-centric analysis of visual attention in MLLMs.} We identify visual attention misalignment as a credit assignment failure and demonstrate that chains-of-thought struggle to recover from early attention errors.
    \item \textbf{A reinforcement learning framework for visual attention learning.} We propose Entropy-Based Target Attention Reward, which introduces an entropy-selective, attention-based reward to enable stable and sustained visual grounding during multimodal reasoning.
    \item \textbf{A strong and generalizable visual reasoning model.} Using the proposed framework, we train SAYO and show consistent improvements across diverse multimodal benchmarks, demonstrating strong generalization in both reasoning and perception tasks.
\end{itemize}
\section{Related Works}
\subsection{MultiModal Large Language Models}
Multimodal Large Language Models ~\citep{step3vl,kimivl}have made remarkable progress by integrating various modalities—such as text, images, and video—into a unified framework for understanding and reasoning. In this framework, different modality encoders project inputs into a shared semantic space, which is then processed by a language model to generate responses. However, although most existing MLLMs possess powerful reasoning capabilities, they struggle to pinpoint the truly relevant parts within complex visual information accurately~\citep{liu2025seeingbelievingprobingdisconnect}. This prevents the reasoning capabilities of MLLMs from being effectively utilized.

\subsection{Enhance Visual Reasoning}
A significant recent trend in research involves explicitly applying visual processing to images using external tools (e.g., Python programs) before inputting them into models, such as resizing images or adding bounding boxes around target objects. Studies indicate that such methods can significantly impact the performance of MLLMs, particularly in tasks like visual localization. Recent studies, such as Visual SketchPad~\citep{visualsketchpad}, ReFocus~\citep{refocus}, and ControlMLLM~\citep{wu2025controlmllmtrainingfreevisualprompt} have explored the performance of visual cueing frameworks across various visual comprehension tasks. However, these methods rely on where MLLMs locate visual prompts for processing. BLINK~\citep{blink} indicates that most open-source multimodal language models struggle to comprehend visual prompts, and incorrect localization undermines their effectiveness. Furthermore, constrained by the fixed nature of the procedures, the aforementioned methods face challenges in transferring to new visual reasoning tasks.

\subsection{Strengthen Visual Focus}
Consistent with our proposed visual focus, recent research~\citep{yang2025learninglookdisentangledcurriculum,zhang2025mllmsknowlooktrainingfree} emphasizes the importance of enhancing attention to visual cues in long-chain reasoning. ~\citep{chen2025perturbollavareducingmultimodalhallucinations} discovered that the model's overreliance on prior language led to neglect of visual details in dense information tasks. Look-Back~\citep{lookback} enhances the focus of the thinking process on images by incorporating look-back labels into long-term reasoning chains. Building upon this foundation, Reflection-V~\citep{lookagain} further introduces an attention reward mechanism. By rewarding the overall attention of the thought chain text toward the image, it promotes the discovery of visual information. Unlike the aforementioned methods, we leverage data annotated with visual bounding boxes to enhance the visual attention capabilities of RL-trained MLLMs. The resulting MLLMs maintain visual focus and reasoning on target objects without requiring lengthy textual inference and reflection processes.
\section{Do MLLMs Know where to focus?}
\label{sec:3}
Recent studies suggest that multimodal large language models (MLLMs) frequently fail to attend to the correct visual regions during reasoning, resulting in inference errors. This observation raises two key questions: (i) whether current MLLMs are able to accurately localize target objects in complex multi-object scenes, and (ii) how visual attention misalignment affects downstream reasoning performance. To answer these questions, we conduct a diagnostic analysis on the GQA~\citep{gqa} dataset, focusing on the relationship between visual attention and model accuracy.


As shown in Figure \ref{fig:first}, attention alignment plays a critical role in inference quality: directing attention to the correct visual regions significantly improves prediction accuracy. To quantitatively characterize this effect, we introduce a target attention ratio, which measures the extent to which a model allocates visual attention to the target object relative to irrelevant image regions. 

We extract attention weights from the final transformer layer, where multimodal fusion is fully realized. Let $\alpha^{(h)}_{t_i,t_g}$ denote the attention weight from generated token $t_g$ to image token $p$ at attention head $h$. We denote by $\mathcal{T_\text{target}}$ the set of image tokens corresponding to the target region and by $\mathcal{T_\text{all}}$ those corresponding to entire image regions. For each generated token $t_g$, we compute the average attention mass assigned to the target and entire image regions, then we average across all attention heads to obtain the overall target attention score $a$ and entire image attention score $v$, respectively:

\begin{equation}
    \label{eq:tar1}
    a
    =
    \frac{1}{H}
    \sum_{h=1}^{H}
    \frac{1}{|\mathcal{T_\text{target}}|}
    \sum_{t_i \in \mathcal{T_\text{target}}} \alpha^{(h)}_{t_i,t_g},
\end{equation}
\begin{equation}
    \label{eq:tar2}
    v
    =
    \frac{1}{H}
    \sum_{h=1}^{H}
    \frac{1}{|\mathcal{T_\text{all}}|}
    \sum_{t_i \in \mathcal{T_\text{all}}} \alpha^{(h)}_{t_i,t_g}.
\end{equation}

Based on these quantities, we define an attention advantage score to quantify visual focus:

\begin{equation}
    \label{eq:tar_main}
    R_a = \frac{1}{2}\left(1 + \tanh\!\left(\log \frac{a+\varepsilon}{v+\varepsilon}\right)\right)
\end{equation}
where $\varepsilon$ is a small constant used for stability.

\begin{figure}
    \centering
    \includegraphics[width=.9\linewidth]{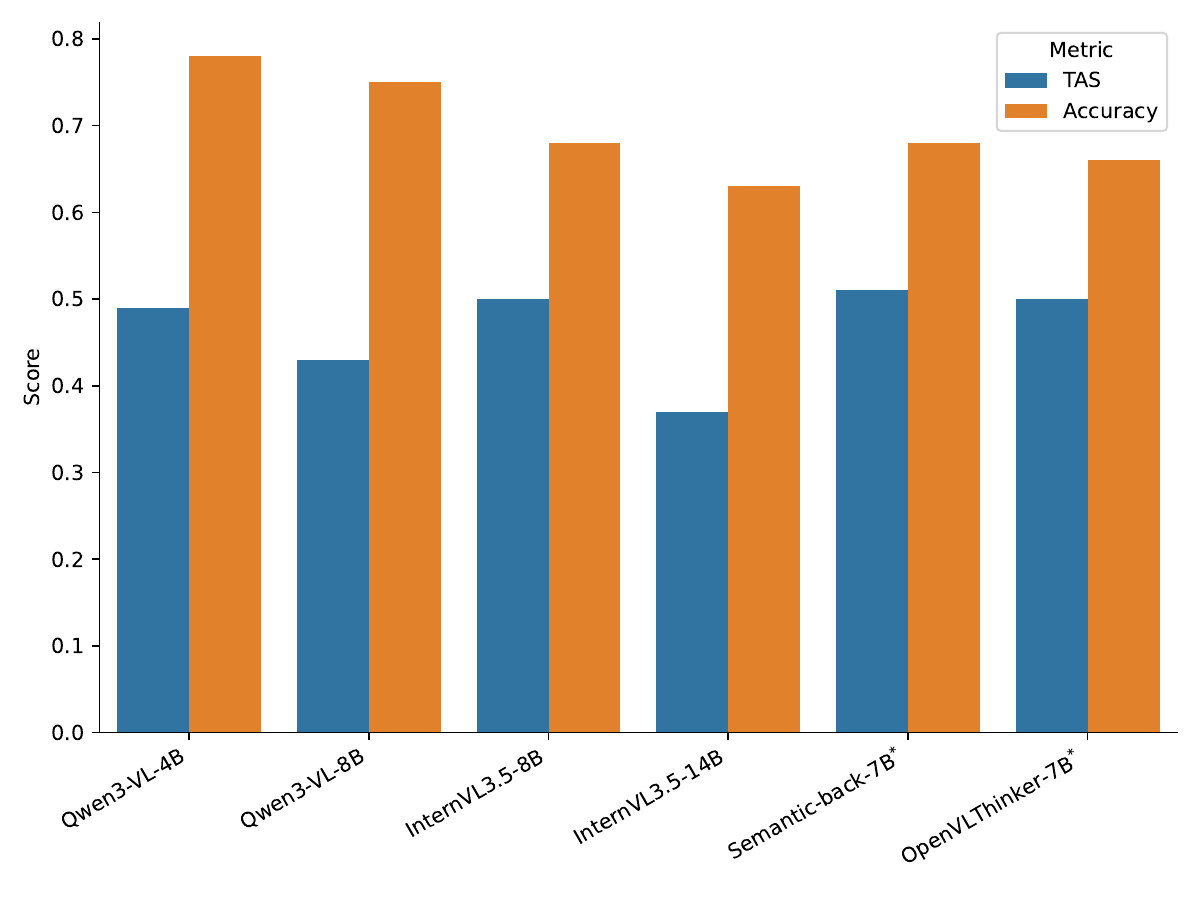}
    \caption{Comparison of target attention score (TAS) and accuracy across Models on a part of GQA dataset. The displayed score and accuracy represent the average across all samples. * denotes models based on the Qwen2.5-7B series.}
    \label{fig:tar_ratio}
\end{figure}


As shown in Figure~\ref{fig:tar_ratio}, experiments across multiple visual models reveal a positive correlation between target attention scores and response accuracy. For models within the same series, higher attention weights on the target visual region yield better inference performance. Notably, despite recent advances in reinforcement learning–based optimization for MLLMs, all evaluated models exhibit consistently low target attention scores. This suggests that while existing RL methods improve textual reasoning trajectories, they fail to provide effective learning signals for precise visual focus. Consequently, models may develop strong reasoning capabilities without reliably grounding their inferences in the correct visual evidence, fundamentally limiting their visual reasoning performance.

To further substantiate the relationship between visual attention and reasoning accuracy, we conducted an additional validation experiment on 100 randomly sampled examples from the GQA dataset. During inference, we progressively reduced the attention weights assigned to answer-relevant visual regions by applying different scaling factors, while keeping the rest of the model unchanged. The results are shown in Table\ref{tab:cmp-tas}:

\begin{table}[t]
    \centering
    \renewcommand{\arraystretch}{1.3}
    \begin{tabular}{l|c}
    \toprule
        Scale Value & Accuracy \\ \midrule
        1.0 (Normal) & 14.0 \\ 
        0.5 & 11.0 \\ 
        0.1  & 10.0 \\ 
        0.0 & 3.0 \\ 
        \bottomrule
    \end{tabular}
    \caption{Comparison of attention weight and accuracy across different scale values on a part of GQA dataset in Qwen3-VL-8B-Instruct.}
    \label{tab:cmp-tas}
\end{table}

These results provide quantitative evidence that weakening the model’s attention to answer-relevant visual regions leads to a clear drop in accuracy.
\section{Method}

\begin{figure*}
    \centering
    \includegraphics[width=.9\textwidth]{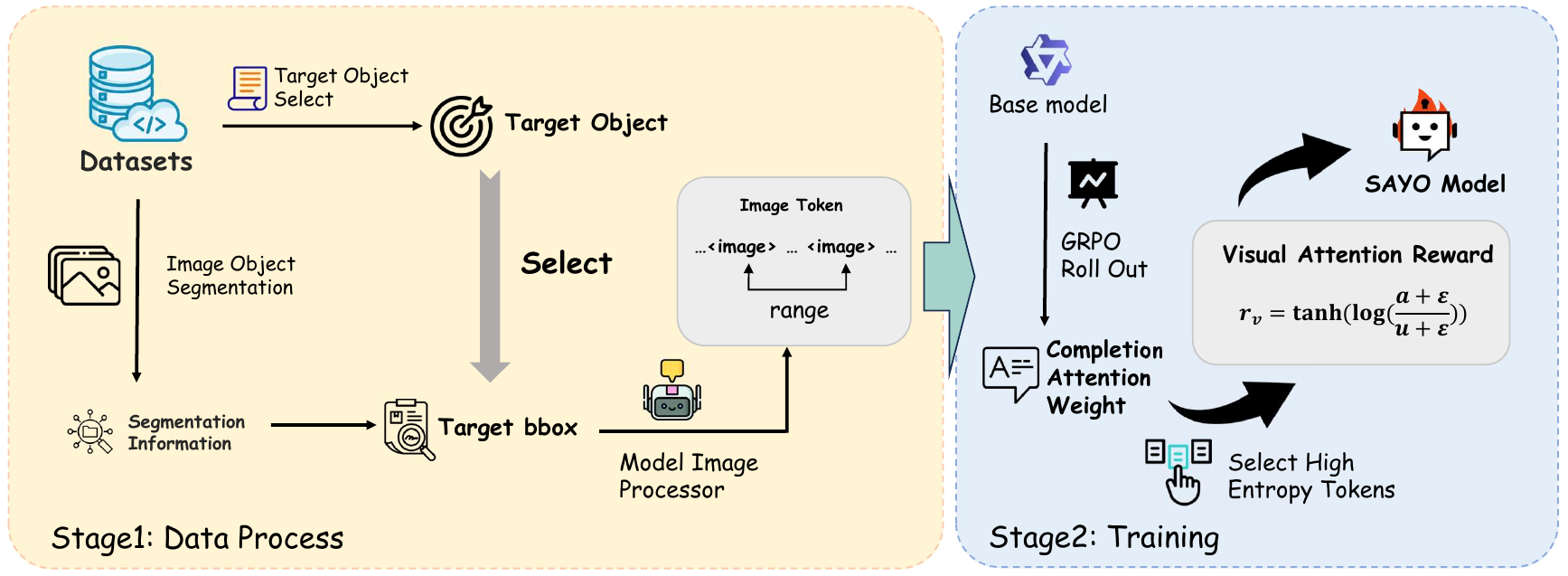}
    \caption{The workflow for our method, including constructing reasoning data with region visual information and training with region attention reward}
    \label{fig:data_process}
\end{figure*}

In the preceding analysis, we observe that current models exhibit insufficient attention to target visual information. This deficiency in visual information utilization severely limits their powerful reasoning capabilities when addressing visual reasoning problems. To address this issue, we propose a reinforcement learning reward training strategy based on visual attention. First, we constructed training data emphasizing visual attention focus using visual reasoning datasets with detailed object annotation information. Subsequently, we employed GRPO to incentivize attention toward target visual signals through reward function.

\subsection{Construction of Data with Visual Focus}
Existing large visual language model training paradigms typically focus solely on the accuracy and formatting compliance of model responses and evaluation accordingly. The neglect of visual attention shifts during reasoning processes prevents these data from demonstrating the visual attention focus we propose. Inspired by recent studies, we employ a precisely designed toolchain that aligns textual questions with visual token information to accomplish data construction tasks. The data construction process is detailed in the following.

As shown in Figure \ref{fig:data_process}, in the first stage, we extract the final target object text from the question-answer pairs and match it with image segmentation information to obtain bounding box coordinates. Subsequently, based on the model's image processor, we convert the bounding boxes into corresponding visual token ranges. This process yields the objects eligible for visual attention rewards.

\subsection{Visual Attention Based Reward}
Following previous studies, we employ the reinforcement learning algorithm GRPO to enhance the perception, localization, and reasoning capabilities of visual models. Building upon the original format reward, we have introduced a new reward mechanism based on target visual attention. This aims to incentivize the model to accurately focus on the correct visual tokens. Based on the analyses in Section \ref{sec:3}, model's attention to target visual regions is positively correlated with accuracy.

Furthermore, prior research~\citep{2080,wang2025sparsemmheadsparsityemerges} has shown that a small subset of high-entropy tokens contains disproportionately rich information. Selecting these information-dense tokens as optimization targets can therefore yield stronger performance. Importantly, in SAYO, high-entropy token selection is used only for reward construction, rather than directly participating in the loss computation. 
We formalize the reasoning process as a trajectory $\tau = (v, q, t_1, t_2, ..., t_T)$. For a high-entropy token $t_k \in \mathcal{Q}{high}$, the model exhibits high epistemic uncertainty, often caused by insufficient grounding in the visual context $v$. Standard Next-Token Prediction minimizes $-\log p(t_k | v, t{<k})$, which allows the model to bypass visual verification and rely instead on linguistic priors, leading to hallucination. In contrast, SAYO does not directly optimize these tokens through the training loss; instead, it uses them to identify uncertain reasoning states for reward assignment. Specifically, we encourage higher visual focus at these high-entropy tokens through the reinforcement learning reward:

\begin{equation}
\mathcal{R}_{t} = \mathbb{I}({t \in Q_{high}}) \cdot r_v(a_t)
\end{equation}

This mechanism effectively suppresses “blind” reasoning. We argue that this “Look-to-Verify” policy constitutes a domain-agnostic meta-skill. Once the model learns to ground its attention in complex natural scenes (dense objects) and charts (structured elements), this attention-sharpening capability naturally transfers to other visual domains, such as geometric diagrams, enabling the pre-trained mathematical reasoning engine to operate on correctly perceived visual primitives.

\begin{table*}[t]
    \centering
    \renewcommand{\arraystretch}{1.26}
    \fontsize{7.2}{7.9}
    \selectfont
    \begin{tabular}{l|cccc|cc|ccc|c}
    \toprule
    \multirow{2}{*}{Model} & \multicolumn{4}{c}{General} & \multicolumn{2}{c}{Math} & \multicolumn{3}{c}{Chart} & \multirow{2}{*}{Avg}\\
    \cline{2-5} \cline{6-7} \cline{8-10} & MMERealWorld & M3CoT & V$^{*}$ & MMStar & MathVision & We-Math & ChartQA & AI2D & CharXiv \\ \midrule
    \multicolumn{11}{c}{\textit{Close-Source Models}} \\ \midrule
    GPT4o & 73.06 & 74.20$^{\dagger}$  & - & 64.70$^{\dagger}$ & 30.40$^{\dagger}$  & 69.00$^{\dagger}$  & 75.32 & 84.60 & 48.90$^{\dagger}$  & -  \\ 
    Gemini 2.5 Pro & - & - & 83.80$^{\dagger}$ & 77.50$^{\dagger}$ &  -  & 80.60$^{\dagger}$  & 83.30$^{\dagger}$ & 90.90$^{\dagger}$ & - & -  \\  \midrule
    \multicolumn{11}{c}{\textit{Open-Source General Models}} \\ \midrule
    Qwen3-VL-4B & 56.80  & 65.10  & 79.06  & 60.80  & 21.64  & 55.63  & 80.12  & 75.16  & 38.00  & 59.15   \\ 
    Qwen3-VL-8B & 56.23  & 64.71  & 81.15  & 62.60  & 22.20  & 52.64  & 78.96  & 75.55  & 42.70  & 59.64   \\ 
    InternVL3\_5-8B & 45.34  & 54.70  & 73.82  & 60.27  & 13.82  & 20.06  & 79.12  & 73.22  & 34.10  & 50.49   \\ 
    InternVL3\_5-14B & 48.57  & 51.86  & 70.16  & 55.93  & 13.85  & 33.10  & 81.76  & 69.43  & 38.90  & 51.51   \\ 
    InternVL3\_5-38B & 54.56  & 53.75  & 79.58  & 54.20  & 13.36  & 24.20  & 81.60  & 75.10  & \textbf{44.40}  & 53.42   \\ 
    Kimi-VL-16B & 42.00  & 54.01  & 75.39  & 63.47  & 17.50  & 42.01  & 82.08  & 77.04  & 31.30  & 53.87   \\ \midrule
    \multicolumn{11}{c}{\textit{Open-Source Reasoning Models}} \\ \midrule
    ViGoRL & 55.34  & \underline{68.72}  & 60.73  & 61.53  & 22.07  & 62.01  & 59.64  & 79.70  & 30.40  & 55.57   \\ 
    OpenVLThinker-7B & 48.31  & 63.29  & 80.10  & 63.07  & 25.59  & 64.48  & 75.44  & \underline{82.61}  & 35.20  & 59.79   \\ 
    Semantic-back-7B & 46.43  & 67.77  & 78.01  & 63.07  & \underline{27.57}  & 66.61  & \textbf{82.32}  & 81.74  & 38.10  & 61.29   \\ 
    NoisyRollout-7B & 51.69  & 67.90  & 76.96  & \underline{63.67}  & 22.01  & \underline{70.57}  & 82.16  & 80.73  & 40.50  & 61.80   \\ \midrule
    \multicolumn{11}{c}{\textit{Ours Trained Model}} \\ \midrule
    SAYO-Qwen-4B & \underline{57.63}  & 64.32  & \textbf{83.25}  & 63.53  & 22.47  & 63.97  & 81.96  & 80.51  & 41.90  & 62.17   \\ 
    SAYO-Qwen-8B & \textbf{65.87}  & \textbf{68.81}  & \underline{82.20}  & \textbf{67.60}  & \textbf{27.99}  & \textbf{73.51}  & 81.16  & \textbf{83.48}  & 42.50  & 65.70   \\ 
    SAYO-Intern-8B & 50.03  & 57.20  & 72.77  & 62.73  & 13.62  & 25.23  & \underline{82.28}  & 76.88  & \underline{43.20}  & 53.77  \\
    \bottomrule
    \end{tabular}
    \caption{Performance of SAYO across various visual reasoning benchmarks with different tasks. $\dagger$ indicates results are taken from the respective models’ official reports. The best results of each benchmark among open-source models are \textbf{bold} and the secondary results are \underline{underlined}.}    
    \label{tab:main}
\end{table*}

Based on these findings, we designed the following reward rules: Let $\mathcal{Q}_{\text{high}}$ denote the set of generated tokens whose entropies rank in the top $30\%$ among all generated tokens. Using attention weights from the final transformer layer~\citep{lookagain}, we compute the average attention mass from the selected tokens to the target visual region and to all visual tokens, respectively:
\begin{equation}
    \label{eq:reward1}
    a_q
    =
    \frac{1}{|\mathcal{Q}_{\text{high}}|}
    \sum_{t_g \in \mathcal{Q}_{\text{high}}}
    \frac{1}{H}
    \sum_{h=1}^{H}
    \frac{1}{|\mathcal{T_\text{target}}|}
    \sum_{t_i \in \mathcal{T_\text{target}}} \alpha^{(h)}_{t_i,t_g},
\end{equation}
\begin{equation}
    \label{eq:reward2}
    v_q
    =
    \frac{1}{|\mathcal{Q}_{\text{high}}|}
    \sum_{t_g \in \mathcal{Q}_{\text{high}}}
    \frac{1}{H}
    \sum_{h=1}^{H}
    \frac{1}{|\mathcal{T_\text{all}}|}
    \sum_{t_i \in \mathcal{T_\text{all}}} \alpha^{(h)}_{t_i,t_g}.
\end{equation}
Therefore, the visual attention-based reward is given by:
\begin{equation}
    \label{eq:reward_main}
    r_v
    =
    \tanh\!\left(
    \log \frac{a_q + \varepsilon}{v_q + \varepsilon}
    \right),
\end{equation}

where $\varepsilon$ is a small constant for numerical stability. The reward $r \in (-1,1)$ reflects whether the model allocates relatively more attention to the target region than to the visual context as a whole. The overall reward $r_o$ in GRPO is the weighted sum of the region visual attention-based reward $r_v$ and format reward $r_f$~\citep{format_reward}, the equation is given as: $r_o = r_v + r_f$

\section{Experiments}

\subsection{Experimental Setup}
\paragraph{Implementations.} To evaluate our method, we use Qwen3-VL~\citep{qwen3vl} and InternVL3.5-8B~\citep{internvl3.5} as the base model. During the training stage, we trained the model using GRPO with entropy and visual attention-based reward. The detailed composition of training data is shown in Appendix \ref{app:c}. Training details are provided in Appendix \ref{app:a}.

\begin{figure*}
    \centering
    \includegraphics[width=.9\textwidth]{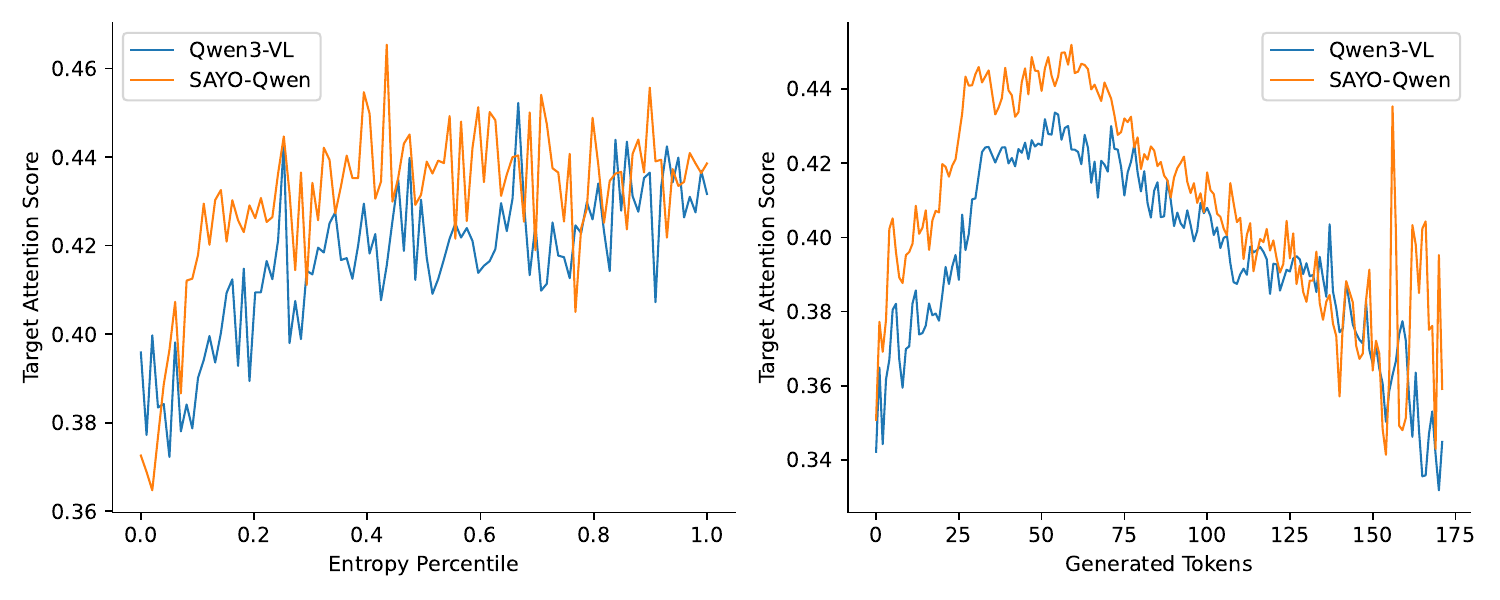}
    \caption{Attention weights of model-generated tokens to target visual tokens (last layer) and attention weights of tokens with different entropy values to target visual tokens. The entropy values shown have been normalized across samples, and the displayed attention weights represent the average across all samples.}
    \label{fig:post_analysis}
\end{figure*}

\paragraph{Benchmarks.} We conducted detailed analytical experiments to evaluate how our approach enhances the model's visual reasoning capabilities. To ensure thoroughness in our assessment, we selected multiple widely recognized visual understanding benchmarks across various domains. These benchmarks encompass structured image reasoning, mathematical reasoning, and general visual reasoning. For evaluating mathematical reasoning, we adopt We-Math~\citep{wemath} and MathVision~\citep{mathvision}. To evaluate performance across general visual reasoning, we utilize M3CoT~\citep{m3cot}, V*Bench~\citep{v*bench}, MMStar~\citep{mmstar}, and MME-RealWorld-Lite~\citep{mmerealworld}. Furthermore, ChartQA~\citep{chartqa}, CharXiv~\citep{chartxiv}, and AI2D~\citep{ai2d} are used to assess structured image reasoning ability, as they cover a broad range of chart understanding questions. Additionally, we compared SAYO against various of baselines: (i) close-source MLLMs such as GPT-4o~\citep{gpt4o} and Gemini 2.5 pro~\citep{gemini2.5}; (ii) open-source general MLLMs, such as Qwen3-VL~\citep{qwen3vl},  InternVL3.5~\citep{internvl3.5}, and Kimi-VL-16B~\citep{kimivl}; (iii) open-source reasoning MLLMs, such as OpenVLThinker-7B~\citep{openvlthinker}, Semantic-back-7B~\citep{lookback}, ViGoRL~\citep{vigorl}, and NoisyRollout-Geo3k-7B~\citep{noisyrollout}.

\subsection{Main Results}
We evaluate the performance of our model, SAYO, on various of visual reasoning benchmarks across three categories: visual math problems, structured image problems, and general reasoning. As shown in Table \ref{tab:main}, the results indicate that our model significantly outperforms base models and other open-source models of a similar scale in terms of reasoning capability. Compared to models trained using other methods, SAYO also demonstrates significant performance advantages. In contrast to existing reasoning MLLMs, which show improved math reasoning but a decline in general reasoning capabilities, SAYO yields significant improvements across a variety of reasoning tasks. Besides, experimental results show that our proposed method is effective across different models and various scales.

\begin{table*}[t]
    \centering
    \renewcommand{\arraystretch}{1.25}
    \fontsize{7.7}{7.8}
    \selectfont
    \begin{tabular}{l|cccc|cc|ccc|c}
    \toprule
    \multirow{2}{*}{Model} & \multicolumn{4}{c}{General} & \multicolumn{2}{c}{Math} & \multicolumn{3}{c}{Chart} & \multirow{2}{*}{Avg}\\
    \cline{2-5} \cline{6-7} \cline{8-10} & MMERealWorld & M3CoT & Vstar & MMStar & MathVision & We-Math & ChartQA & AI2D & CharXiv \\ \midrule
    Qwen3-VL-8B & 56.23  & 64.71  & 81.15  & 62.60  & 22.20  & 52.64  & 78.96  & 75.55  & 42.70  & 59.64   \\ 
    + Attn. Reward & 62.85  & 67.86  & 82.20  & 65.27  & 25.26  & 64.83  & 81.84  & 83.06  & 42.50  & 63.96   \\ 
    + Acc. Reward & 57.01  & 65.96  & 78.01  & 65.20  & 23.59  & 56.61  & 82.52  & 78.01  & 41.40  & 60.92   \\ 
    Acc. + Attn. & 61.59  & 67.77  & 82.20  & 66.00  & 25.95  & 66.15  & 82.56  & 83.03  & 43.20  & 64.27   \\ \midrule
    InternVL3.5-8B & 45.34  & 54.70  & 73.82  & 60.27  & 13.82  & 20.06  & 79.12  & 73.22  & 34.10  & 50.49   \\ 
    + Attn. Reward & 50.03  & 57.20  & 72.77  & 62.73  & 13.62  & 25.23  & 82.28  & 76.88  & 43.20  & 53.77   \\ 
    + Acc. Reward & 47.94  & 57.16  & 74.35  & 62.20  & 14.05  & 24.31  & 82.68  & 75.62  & 41.60  & 53.32   \\ 
    Acc. + Attn. & 49.04  & 56.77  & 74.35  & 63.73  & 14.14  & 27.24  & 82.04  & 76.81  & 42.80  & 54.10  \\
    \bottomrule
    \end{tabular}
    \caption{Ablation results for different combinations of rewards and visual attention-based reward on performance improvement. Attn. Reward denotes visual attention-based reward, and Acc. Reward denotes accuracy reward. All combinations include format rewards.}
    \label{tab:ablation_reward}
\end{table*}

A counter-intuitive finding in Table \ref{tab:main} is the significant improvement on We-Math and MathVision, despite the exclusion of mathematical datasets from our training phase. We attribute this cross-domain generalization to the decoupling of visual parsing and logical reasoning. Current state-of-the-art base models already possess strong latent mathematical reasoning capabilities derived from massive textual pre-training. However, their performance is often bottlenecked by visual misalignment which propagates errors into the reasoning chain. By training with visual attention focus, SAYO learns a robust structure-aware attention policy to correcting the input signal and effectively eliminating the garbage in phase. This allows the model's pre-existing mathematical engine to process valid visual premises, thereby unlocking performance gains without explicit domain-specific training.

\subsection{Ablation Study}
\label{sec:5-3}
\paragraph{Effectiveness of attention reward.} To systematically assess the contribution of the proposed regional visual attention–based reward, we conduct a series of controlled ablation experiments designed to isolate its effect from other reward components. Specifically, we consider the following variants:
(i) Accuracy-only reward, where the attention-based reward is replaced by a conventional answer accuracy reward;
(ii) Attention-only reward, where optimization relies solely on the proposed visual attention reward;
(iii) Combined reward, where both attention-based and accuracy rewards are applied.

The results, summarized in Table~\ref{tab:ablation_reward}, demonstrate that incorporating a visual attention–based reward leads to substantial and consistent performance improvements across benchmarks. Notably, models trained using only visual attention reward achieve performance comparable to those trained with the combined rewards, whereas models trained with accuracy reward alone exhibit only marginal gains.

These findings suggest that deficiencies in current MLLMs stem less from limited reasoning capacity and more from insufficient visual perception and localization. Basically, while existing models are capable of performing complex reasoning once relevant information is identified, they often fail to reliably extract and attend to the necessary visual evidence. By explicitly incentivizing correct visual focus, the proposed attention reward effectively unlocks the model’s latent reasoning capabilities, leading to more accurate and robust inference.

\begin{table*}[t]
    \centering
    \renewcommand{\arraystretch}{1.25}
    \fontsize{7.7}{7.8}
    \selectfont
    \begin{tabular}{l|cccc|cc|ccc|c}
    \toprule
    \multirow{2}{*}{Model} & \multicolumn{4}{c}{General} & \multicolumn{2}{c}{Math} & \multicolumn{3}{c}{Chart} & \multirow{2}{*}{Avg}\\
    \cline{2-5} \cline{6-7} \cline{8-10} & MMERealWorld & M3CoT & Vstar & MMStar & MathVision & We-Math & ChartQA & AI2D & CharXiv \\ \midrule
    Qwen3-VL-8B & 56.23  & 64.71  & 81.15  & 62.60  & 22.20  & 52.64  & 78.96  & 75.55  & 42.70  & 59.64   \\ 
    key tokens & 62.85  & 67.86  & 82.20  & 65.27  & 25.26  & 64.83  & 81.84  & 83.06  & 42.50  & 63.96   \\ 
    all tokens & 56.59  & 66.01  & 75.92  & 64.07  & 22.57  & 58.39  & 82.00  & 77.43  & 41.60  & 60.51   \\ \midrule
    InternVL3.5-8B & 45.34  & 54.70  & 73.82  & 60.27  & 13.82  & 20.06  & 79.12  & 73.22  & 34.10  & 50.49   \\ 
    key tokens & 50.03  & 57.20  & 72.77  & 62.73  & 13.62  & 25.23  & 82.28  & 76.88  & 43.20  & 53.77   \\ 
    all tokens & 46.64  & 55.22  & 72.77  & 63.40  & 13.75  & 24.20  & 81.80  & 74.79  & 41.90  & 52.72  \\
    \bottomrule
    \end{tabular}
    \caption{Comparative results of visual attention reward for all generated tokens and key generated tokens on reasoning performance improvement.}
    \label{tab:ablation_ways}
\end{table*}
\paragraph{Design an Effective Attention Reward} Beyond validating the necessity of the attention-based reward, we further investigate how different design choices affect its effectiveness. We evaluate several reward configurations that vary in reward token selection strategy, as reported in Table~\ref{tab:ablation_ways}. Across all settings, introducing visual attention–based supervision consistently improves performance, confirming the general effectiveness of this approach.

\begin{figure*}
    \centering
    \includegraphics[width=.9\textwidth]{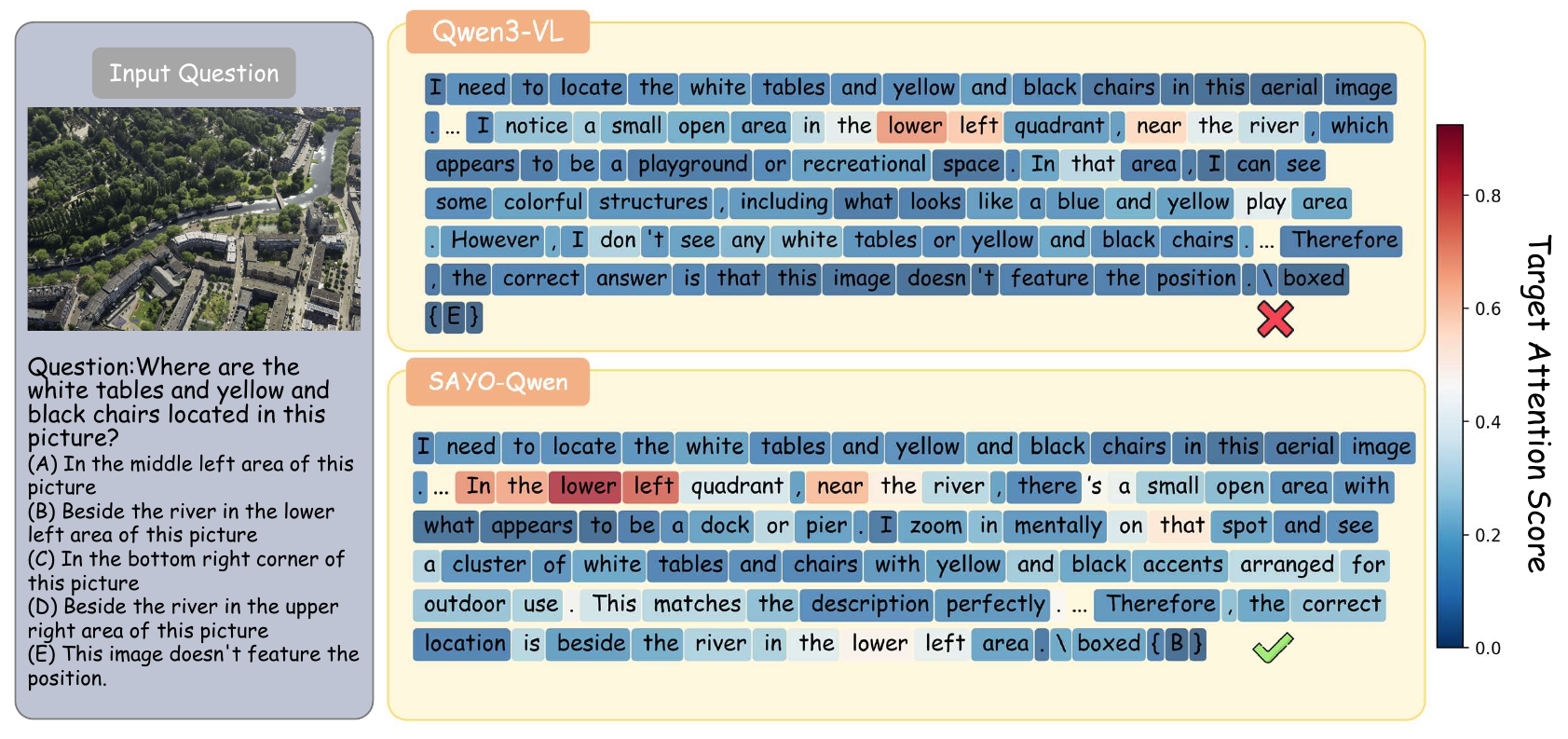}
    \caption{An example demonstrating how areas of visual attention shift during the reasoning process. The background color of tokens in the figure indicates the magnitude of the target visual attention score.}
    \label{fig:case_study}
\end{figure*}

A key observation is that selectively rewarding a subset of high-information tokens yields significantly larger performance gains than applying rewards across all tokens. Further analysis of the training process reveals that entropy-selective token rewards lead to more stable and faster convergence, as evidenced by reduced variance in training rewards and improved consistency. By focusing the reward on tokens that correspond to critical decision points, model receives clearer supervision. Additional analyses of training stability and behavior are provided in Appendix \ref{app:d} and \ref{app:e}.

\subsection{Results Analysis}
In Section \ref{sec:3}, we demonstrate that existing MLLMs exhibit shortcomings in terms of accurate visual perception capabilities through attention advantage score $R_a$. Based on this metric, we will analyze in this section whether the performance improvement of SAYO truly stems from the training strategy targeting regional visual attention, and what changes this training strategy brings to the model.

We compared the target visual attention score $R_a$ between Qwen3-VL-8B-Instruct and SAYO-Qwen-8B on the same task. As shown in Figure \ref{fig:post_analysis}, we found that throughout the generated token sequence, SAYO consistently exhibited significantly higher visual attention weights toward the target region than Qwen3-VL, and significantly enhanced attention to the target region of the image during the later stages of inference when generating answers. Additionally, to verify whether this method can effectively improve the attention of high-entropy tokens toward target visual regions, we compared the attention weights assigned to these regions by SAYO and Qwen3-VL when generating sequences with tokens of varying entropy levels. As shown in Figure \ref{fig:post_analysis}, compared to the baseline model, SAYO maintains consistently high visual attention on most tokens with higher entropy.

As discussed in previous sections, model itself possesses sufficient reasoning capabilities, with performance limitations primarily stemming from deficiencies in focusing on relevant visual information. Our method enhances reasoning performance by strengthening the model's attention to effective visual information. Figure \ref{fig:case_study} presents a comprehensive example demonstrating that these gains are indeed due to the model’s visual focus ability. In this example, our model correctly identified and focused on the object mentioned in the problem during the early stages of reasoning, and consistently maintained high visual attention on that region at critical junctures of the reasoning process. This demonstrates the key mechanism for visual reasoning. As shown in Figure \ref{fig:case_study}, sustained and accurate visual attention can continuously guide the reasoning process away from erroneous visual information, thereby arriving at the correct conclusion. In summary, our method demonstrates a more precise focus on visual information.
\section{Conclusion}
In this paper, we propose that the key factor influencing the reasoning performance of visual models is their ability to perceive and focus on critical visual information. Through quantitative research, we reveal that existing models exhibit poor attention to visual regions containing critical information for inference. We analyze the relationship between visual attention and inference accuracy: visual attention significantly impacts a model's accuracy. To address this bottleneck, we propose a region-level visual attention enhancement training strategy: integrating token-level visual attention rewards into RL training. This strategy enhances model reasoning performance across multiple benchmarks, validating that precise visual attention effectively improves model capabilities.
\section*{Limitations}
While our method significantly outperforms other methods under multiple tasks, the performance gains are different among various models, especially the weaker models gain less than the stronger models. This discrepancy suggests that benefits of visual attention reward are more pronounced in models with a higher baseline capacity, highlighting its potential to enhance the performance of more powerful architectures more effectively. Moving forward, we plan to explore strategies for improving omni modal applicability, aiming to achieve more excellent performance across a broader range of modality.

\section*{Ethic Consideration}
Our training data is sampled from open-source datasets, and the training framework is open-source code, edited by our own proprietary code. All models used are open-source, and their sources are clearly credited. The entire process follows transparent and ethical guidelines, ensuring there are no ethical concerns or issues with the data generation. We are committed to maintaining high standards of integrity and transparency in our work.

\bibliography{content/ref}

\newpage
\appendix
\section{Implementation of Training Details and Hyperparameters}
\label{app:a}

During the Group Relative Policy Optimization (GRPO) phase, we limit max responses to 110 tokens and apply KL divergence with a coefficient of $1e^{-3}$. We use 6 GPUs with 4 epochs of training on TRL~\citep{trl} framework. More details and hyperparameters are shown in Table ~\ref{tr_detail}
\begin{table}[h]
    \centering
    \begin{tabular}{cc}
    \toprule
        Parameters & Value \\ \midrule
        Epochs & 4 \\ 
        Per Device Batch Size & 64 \\ 
        Warmup & False \\ 
        Rollout & 16 \\ 
        Rollout Temperature & 1.0 \\ 
        Rollout Top-P & 0.9 \\ 
        KL divergence coefficient & $1\times10^{-3}$ \\ 
        Learning rate & $5\times10^{-6}$ \\
        Weight Decay & $1\times10^{-2}$ \\
        Max Grad Norm & 0.8 \\
        Optimizer & AdamW \\
    \bottomrule
    \end{tabular}
    \caption{The hyperparameters used during GRPO with visual attention based reward.}
    \label{tr_detail}
\end{table}
\section{Prompts in Training and Evaluation}
\label{app:b}
\begin{tcolorbox}[title=Prompt Template in GRPO]
    A conversation between a user and an assistant. The user asks a question, and the assistant must answer it.
    
    The assistant first conducts internal reasoning, followed immediately by the final answer enclosed within \textless{}answer\textgreater{}...\textless{}/answer\textgreater{} tags.
    
    Requirements:
    
    1. The reasoning outside \textless{}answer\textgreater{}\textless{}/answer\textgreater{} should be concise, minimal, and essential.
    
    2. The final answer must be correct, complete, and directly address the user’s question.

    \textless{}question\textgreater{}
\end{tcolorbox}

\begin{tcolorbox}[title=Prompt Template in Evaluation]
    You FIRST think about the reasoning process as an internal monologue and then provide the final answer. The reasoning process MUST BE enclosed within \textless{}think\textgreater{} \textless{}/think\textgreater{} tags. The final answer MUST BE put in \\boxed\{\}.
\end{tcolorbox}

\begin{tcolorbox}[title=Prompt Template using for ablation study]
    A conversation between a user and an assistant. The user asks a question, and the assistant must answer it.
    
    The assistant first conducts internal reasoning, followed immediately by the final answer enclosed within \textless{}answer\textgreater{}...\textless{}/answer\textgreater{} tags.
    
    Requirements:
    
    1. The reasoning outside \textless{}answer\textgreater{}\textless{}/answer\textgreater{} should be concise, minimal, and essential.
    
    2. The final answer must be correct, complete, and directly address the user’s question.

    3. When reasoning, focus on the areas of aim object in the image.

    \textless{}question\textgreater{}
\end{tcolorbox}
\section{Training Data Source}
\label{app:c}
We collected data from the dataset in the field of real-world visual reasoning and the structured graph reasoning for RL training. The specific composition is shown in Table \ref{tab:data_composition}.

\begin{table}[h]
    \centering
    \begin{tabular}{lc} \toprule
    Datasets & Size \\ \midrule
    GQA~\citep{gqa} & $\sim$ 16k \\
    Refocus\_{}Data~\citep{refocus} & $\sim$ 4k \\ \bottomrule
    \end{tabular}
    \caption{Detailed composition of the datasets used for RL training stage.}
    \label{tab:data_composition}
\end{table}
\section{Experiments Results}
\label{app:d}

\begin{figure*}
    \centering
    \includegraphics[width=0.9\textwidth]{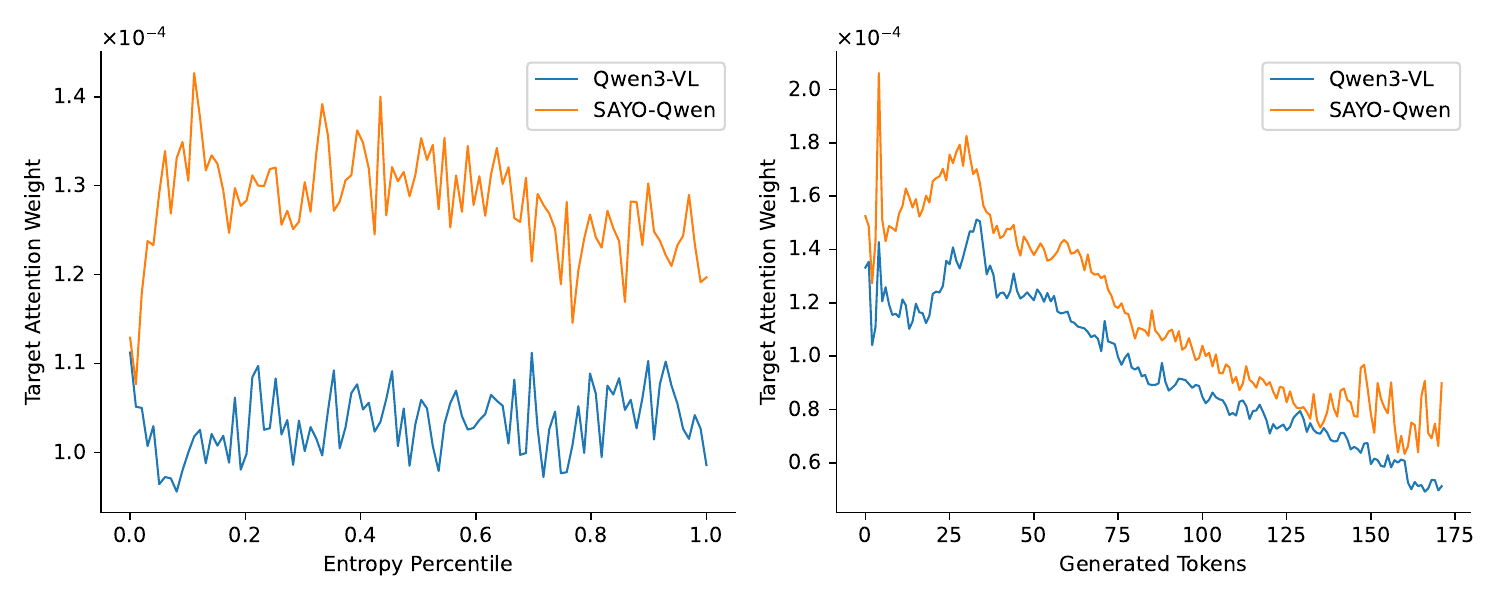}
    \caption{Attention weight (last layer) on target visual token in Qwen3-VL-8B-Instruct and SAYO-8B. The entropy
values shown have been normalized across samples, and the displayed attention weights represent the average across
all samples.}
    \label{fig:taw}
\end{figure*}

\begin{figure*}
    \centering
    \includegraphics[width=.9\textwidth]{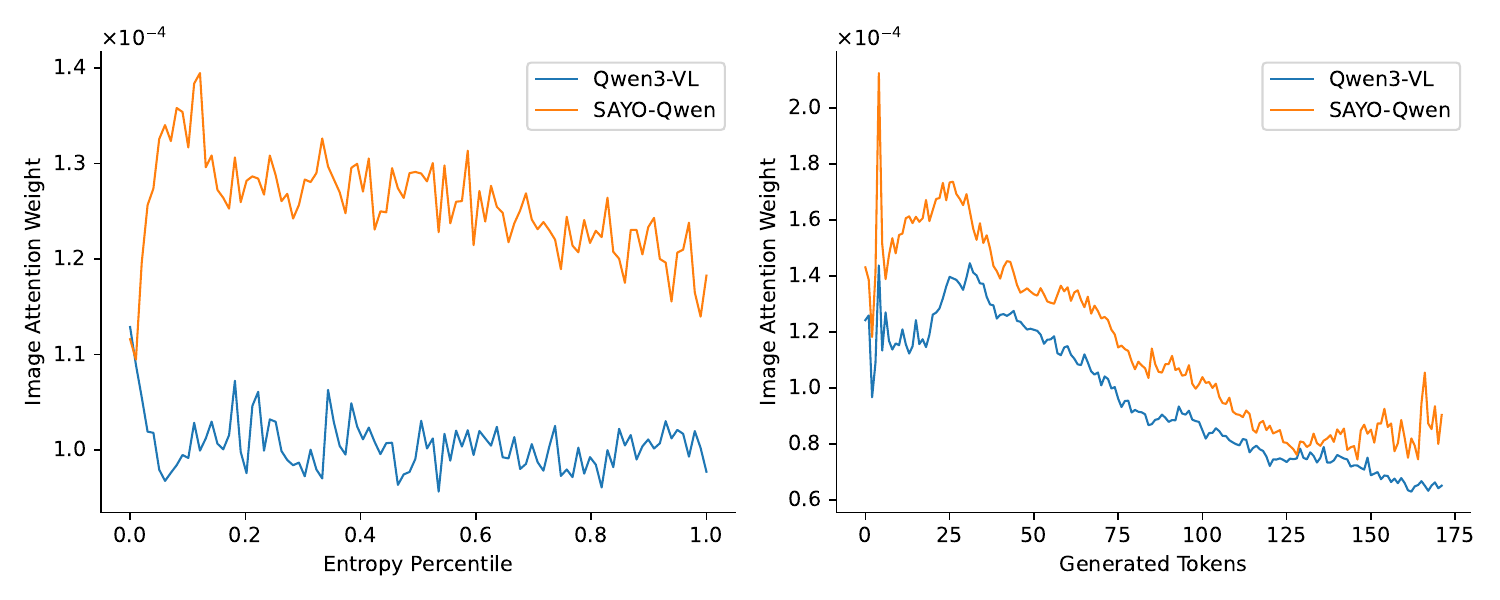}
    \caption{Attention weight (last layer) on all visual tokens in Qwen3-VL-8B-Instruct and SAYO-8B. The entropy values shown have been normalized across samples, and the displayed attention weights represent the average across all samples.}
    \label{fig:iaw}
\end{figure*}

\begin{table*}[t]
    \centering
    \renewcommand{\arraystretch}{1.3}
    \begin{tabular}{l|cc|cc|cc}
    \toprule
    \multirow{2}{*}{Range} & \multicolumn{2}{c}{General} & \multicolumn{2}{c}{Math} & \multicolumn{2}{c}{Chart}\\
    \cline{2-3} \cline{4-5} \cline{6-7} & MMERealWorld & M3CoT & MathVision & We-Math & AI2D & CharXiv \\ \midrule
    Sayo-Qwen-8B & 62.85  & 68.46  & 25.26  & 64.83  & 83.06  & 42.50  \\ 
    Top 20\% tokens & 58.36 & 65.19 & 17.80 & 59.08 & 79.40 & 40.90 \\ 
    Top 40\% tokens & 62.48 & 66.18 & 20.13 & 66.84 & 82.93 & 39.30 \\
    \bottomrule
    \end{tabular}
    \caption{Ablation results for different ranges of entropy select strategies in visual attention-based reward.}
    \label{tab:ablation_entropy}
\end{table*}

\begin{table*}[t]
    \centering
    \renewcommand{\arraystretch}{1.3}
    \begin{tabular}{l|ccc}
    \toprule
    Model & MMIQ & ScienceQA & MMMU-Pro \\ \midrule
    Qwen3-VL-8B & 16.19 & 88.27 & 42.41 \\
    Sayo-Qwen-8B & 19.93 & 89.46 & 45.61 \\
    \bottomrule
    \end{tabular}
    \caption{Extend evaluation results on abstract reasoning tasks.}
    \label{tab:extra_abstract}
\end{table*}

\paragraph{Detailed Study of Visual Attention Shifts} We further evaluated SAYO's optimization of visual attention on the test dataset. As shown in Figures\ref{fig:taw} and \ref{fig:iaw}, SAYO maintains higher attention to both the target region and the entire image throughout the generation process compared to baseline models. When comparing tokens with different entropies, SAYO significantly enhances visual attention for all tokens except those with extremely low entropy.

\paragraph{Detailed Study of Entropy Sensitivity} The result in \ref{sec:5-3} aligns with the intuition that not all tokens contribute equally to visual grounding during multimodal reasoning. Many tokens serve primarily syntactic or connective roles and do not require direct visual attention. Including such low-information tokens in reward computation introduces noise, weakening the learning signal and potentially destabilizing policy optimization. To further investigate the impact of entropy selection range on attention rewards, we trained multiple entropy-based reward token selection ranges. As shown in Table \ref{tab:ablation_entropy}, selecting fewer tokens will exclude some tokens carrying important information, leading to reduced training effectiveness. Conversely, selecting more tokens introduces additional training noise, thereby diminishing training outcomes.

\begin{figure*}
    \centering
    \includegraphics[width=\textwidth]{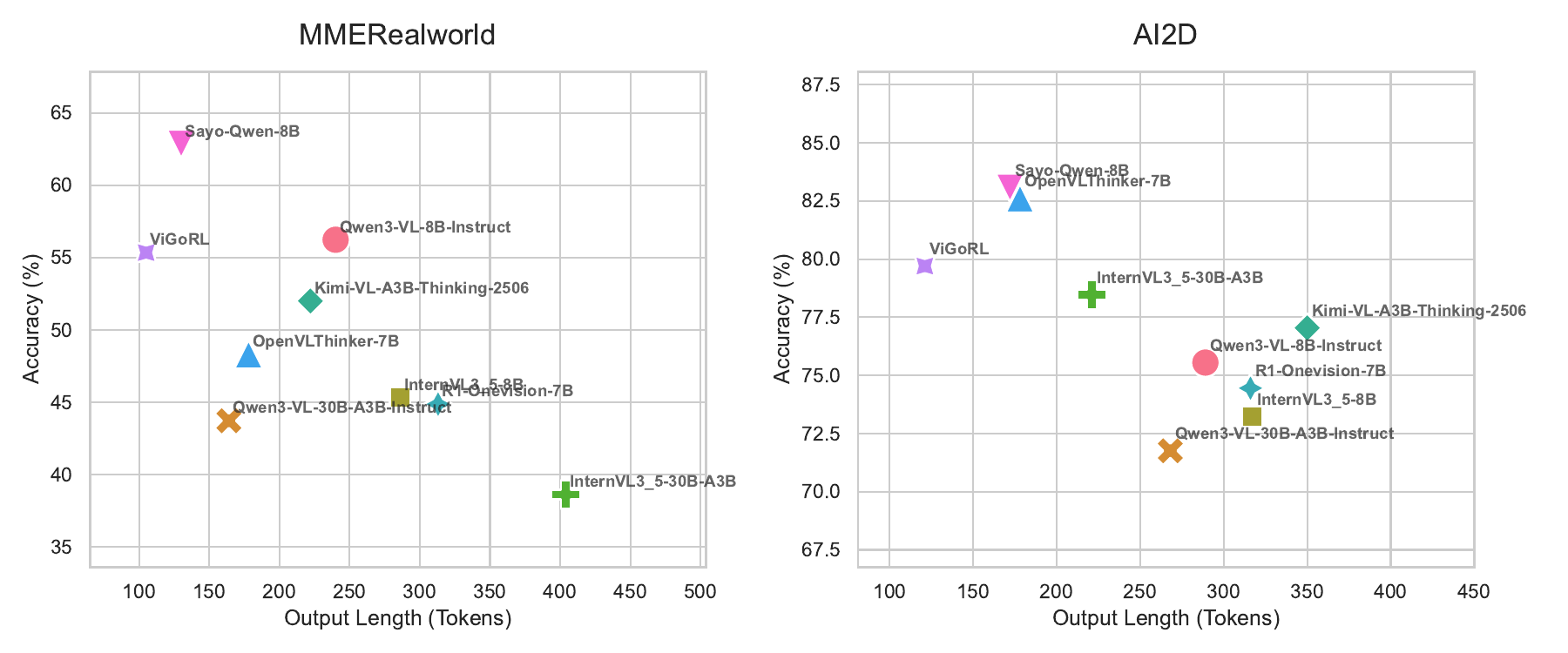}
    \caption{Comparative analysis of model accuracy and average output length across MMERealworld and AI2D.}
    \label{fig:len_comp}
\end{figure*}

\paragraph{Improvement in Model Inference Length} Figure \ref{fig:len_comp} illustrates the relationship between model output length and accuracy across different benchmarks. Thanks to improved visual attention, SAYO achieves higher reasoning performance with shorter reasoning lengths. This demonstrates that accurate visual perception can effectively reduce the extra inference required by the model due to ambiguous visual information.

\paragraph{Evaluation on Abstract Tasks and Multidisciplinary Tasks} To more broadly evaluate the performance of our model on abstract reasoning tasks and multidisciplinary tasks, we conducted evaluation experiments on ScienceQA~\citep{scienceqa}, MMIQ~\citep{mmiq} and MMMU-Pro~\citep{yue-etal-2025-mmmu} benchmarks. Table \ref{tab:extra_abstract} demonstrates that our method can effectively enhance the model’s reasoning capabilities even for tasks without explicit visual targets, highlighting its strong generalization ability.

\paragraph{Ablation Study of Layer Difference} We attempted to train visual attention rewards at different layers; the evaluation results on some benchmarks are as Table~\ref{tab:ex-ld}:

\begin{table}[t]
    \centering
    \renewcommand{\arraystretch}{1.3}
    \begin{tabular}{l|cc}
     \toprule
        Target Layers & MMMU\_Pro & MMstar  \\ \midrule
        10 & 42.45 & 65.60  \\ 
        20 & 41.92 & 63.27  \\ 
        36 (last) & 45.61 & 67.60 \\ \bottomrule
    \end{tabular}
    \caption{Ablation experiments on model's different layer.}
    \label{tab:ex-ld}
\end{table}

The results show that training performance on the last layer is better than on the other layers. While Figure \ref{fig:tar_ratio} presents a cross-model observation to motivate our hypothesis, the above intervention experiment provides complementary within-model evidence by modifying the attention distribution of the same model while keeping all other factors fixed.

\section{Training Behaviors}
\label{app:e}
\begin{figure*}
    \centering
    \includegraphics[width=.9\textwidth]{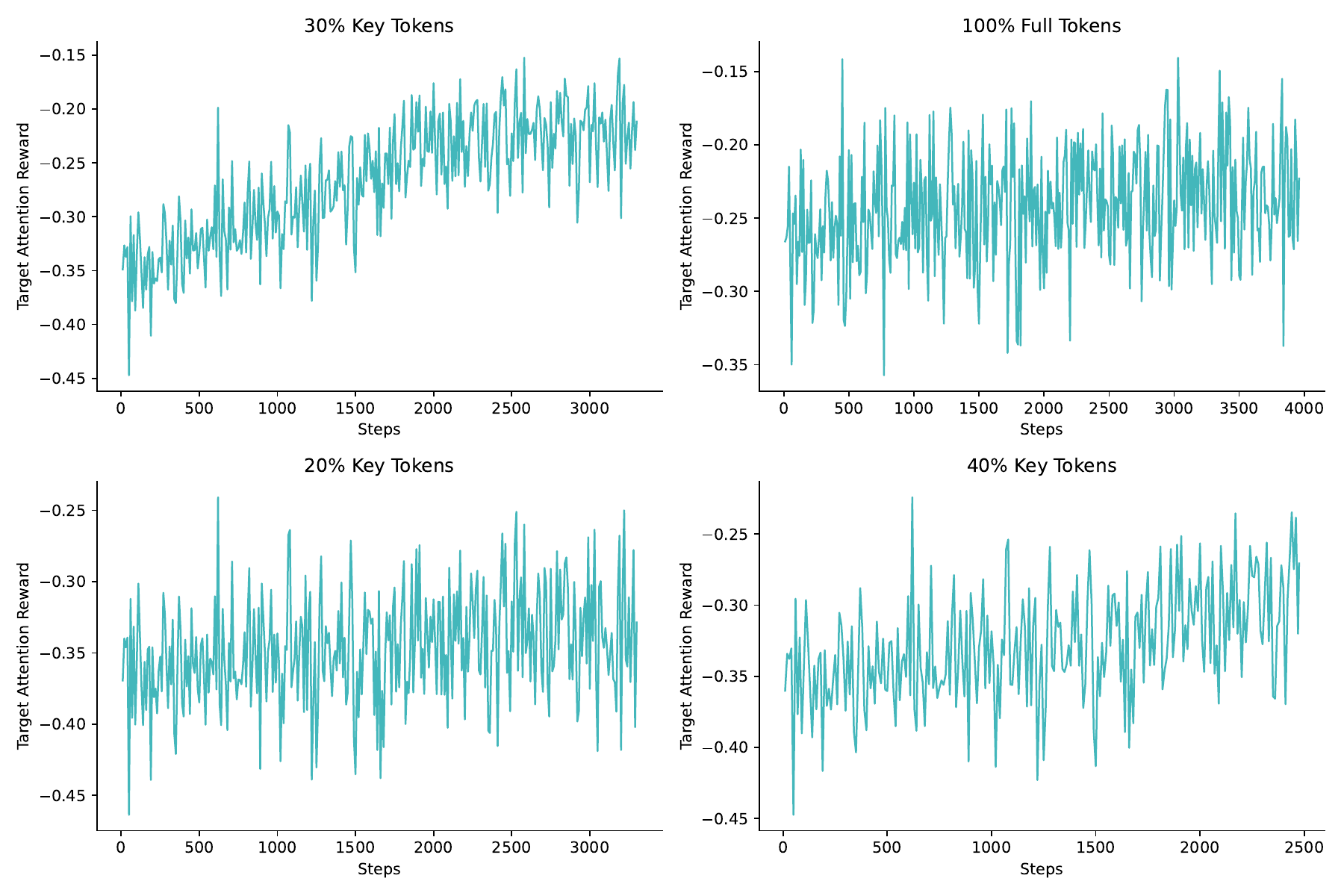}
    \caption{Attention rewards vary with training step. The left figure shows rewards covering the top 30\% of high-entropy tokens, while the right figure covers all tokens.}
    \label{fig:cpt1}
\end{figure*}

Figure \ref{fig:cpt1} shows the training details for selecting top token rewards versus full token rewards. After excluding low-entropy tokens, the reward values can increase normally during training due to reduced noise signals. However, if the range of tokens selected for rewards is too narrow, it will actually increase training instability. We speculate that an overly narrow selection range prevents some tokens containing critical information from receiving rewards, thereby destabilizing the training process. In addition, reducing the range of choices lowers the computational cost of calculating visual attention weights and increases training speed.

\section{Case Study}
\label{app:f}

\begin{figure*}
    \centering
    \includegraphics[width=\linewidth]{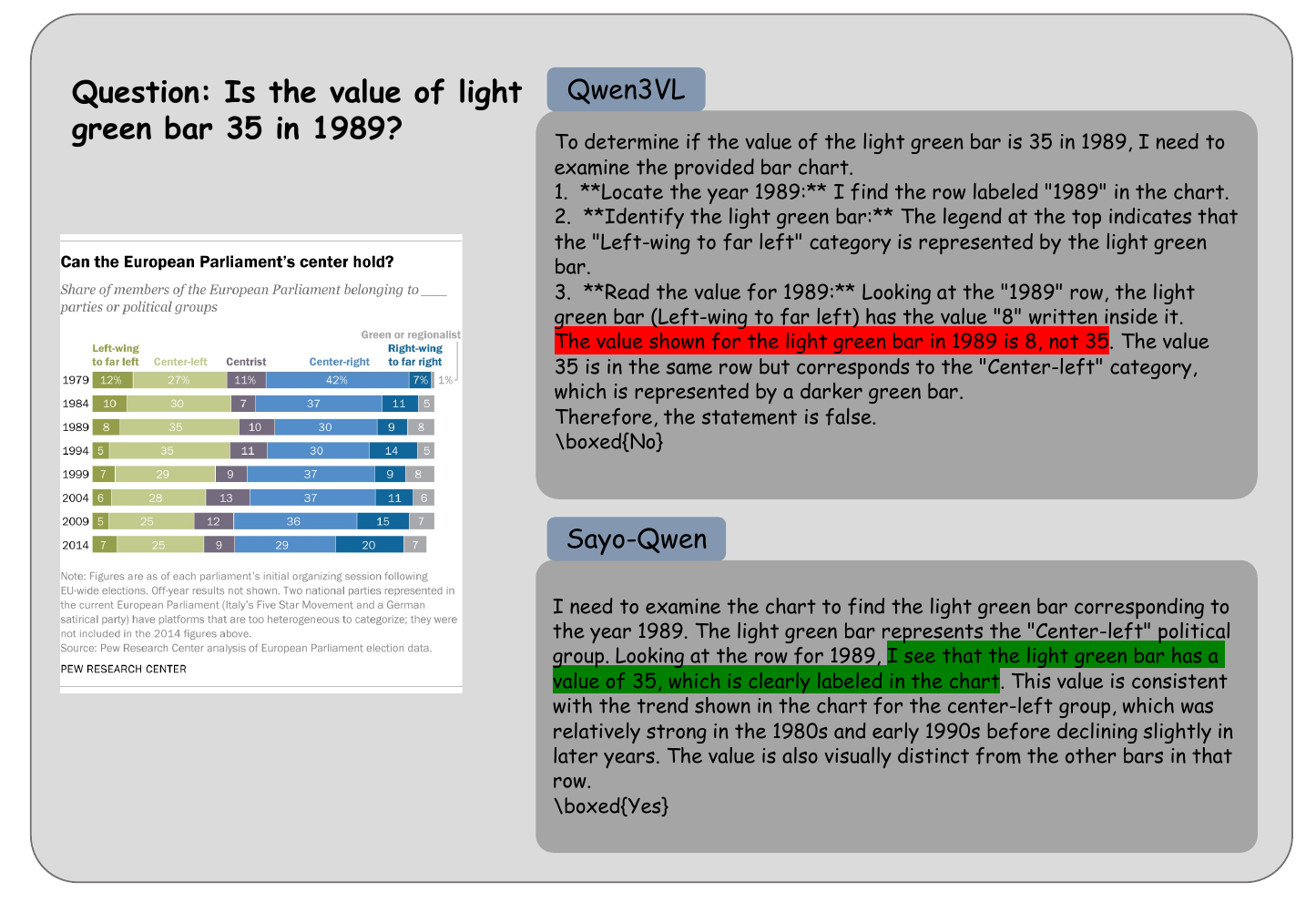}
    \caption{Case 1 of Sayo-Qwen-8B in ChartQA.}
    \label{fig:case1}
\end{figure*}

\begin{figure*}
    \centering
    \includegraphics[width=\linewidth]{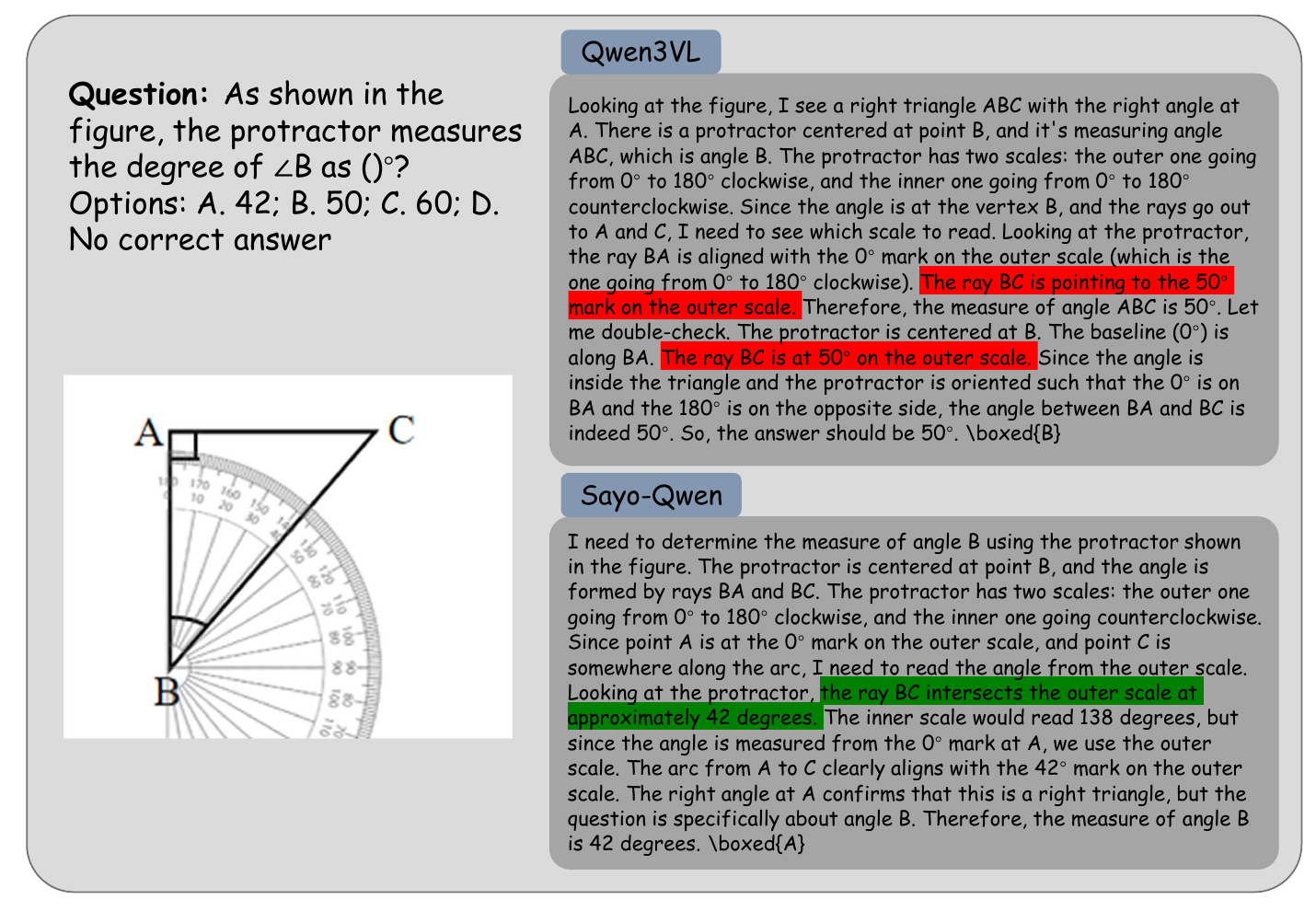}
    \caption{Case 2 of Sayo-Qwen-8B in We-Math}
    \label{fig:case2}
\end{figure*}

\begin{figure*}
    \centering
    \includegraphics[width=\linewidth]{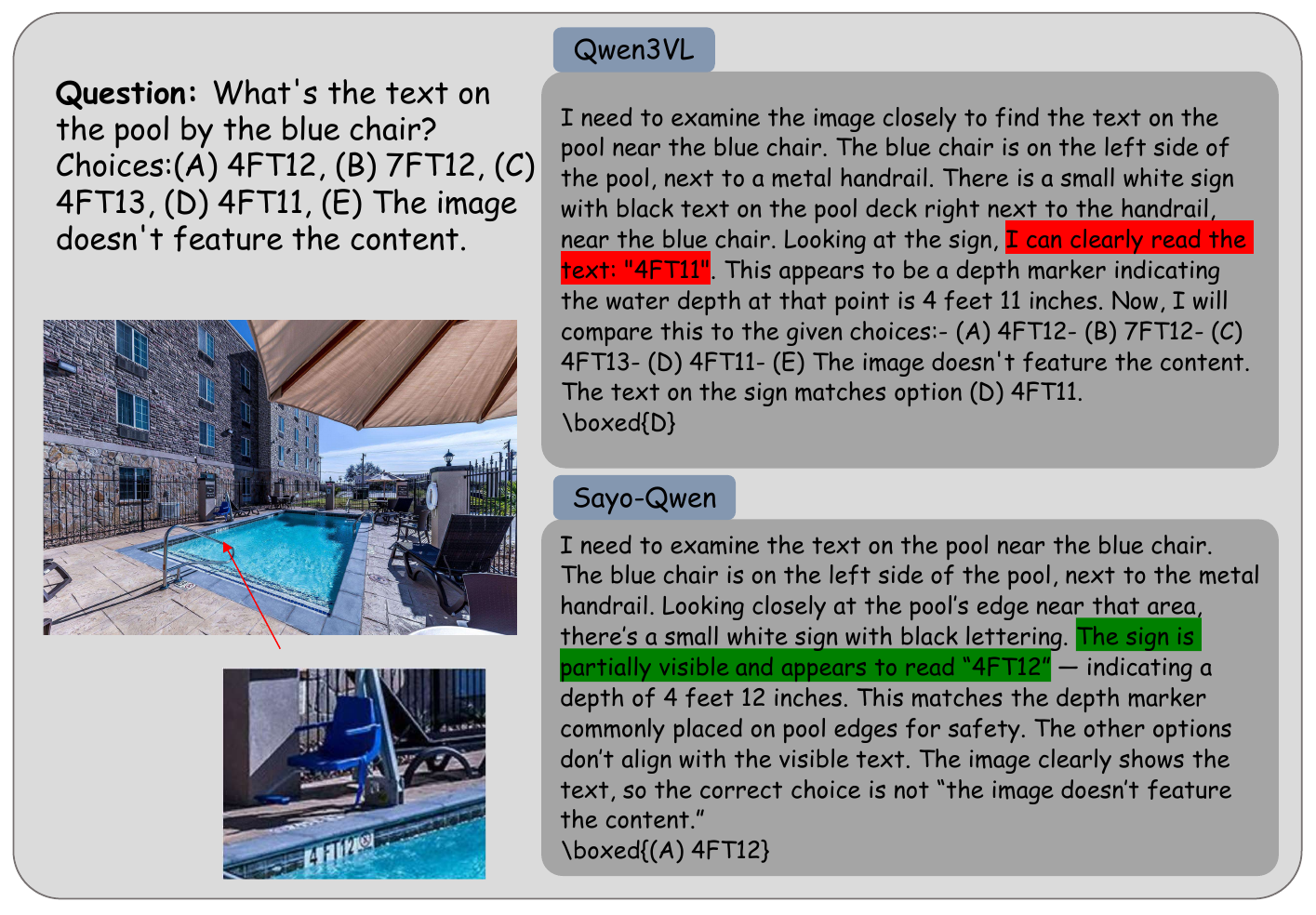}
    \caption{Case 3 of Sayo-Qwen-8B in MME-Realworld}
    \label{fig:case3}
\end{figure*}

Figure \ref{fig:case1}, \ref{fig:case2}, \ref{fig:case3} show examples of performance on benchmarks in different domains. The erroneous visual information obtained by the baseline model significantly affects the model’s performance during inference, whereas SAYO ensures the accuracy of inference by improving the model’s perception of visual information.

\end{document}